\documentclass{article}

\PassOptionsToPackage{numbers, compress}{natbib}

\usepackage[final]{neurips_2020_ml4ad} 

\usepackage[utf8]{inputenc} 
\usepackage[T1]{fontenc}    
\usepackage{url}
\usepackage{hyperref}
\usepackage{xcolor}
\hypersetup{
    colorlinks = true,
    allcolors  = blue, 
}

\usepackage{booktabs}       
\usepackage{amsfonts}       
\usepackage{nicefrac}       
\usepackage{microtype}      

\usepackage{epsfig}
\usepackage{graphicx}

\usepackage{mathptmx} 
\usepackage{times} 
\usepackage{amsmath} 
\usepackage{amssymb}  
\usepackage{amsmath,amssymb,amsfonts}
\usepackage{textcomp}
\usepackage{algorithm}
\usepackage{algpseudocode}
\usepackage{cite}
\usepackage{multirow}

\title{{\fontsize{16.3}{2}\selectfont DeepSeqSLAM: A Trainable CNN+RNN for Joint Global Description and Sequence-based Place Recognition}}

\author{%
  Marvin Chanc\'an \\
  QUT, Australia\\
  \texttt{mchancanl@uni.pe} \\
  \And
  Michael Milford \\
  QUT, Australia \\
  \texttt{michael.milford@qut.edu.au} \\
}

\begin{document}

\maketitle

\begin{abstract}
    Sequence-based place recognition methods for all-weather navigation are well-known for producing state-of-the-art results under challenging day-night or summer-winter transitions. These systems, however, rely on complex handcrafted heuristics for sequential matching---which are applied on top of a pre-computed pairwise similarity matrix between reference and query image sequences of a single route---to further reduce false-positive rates compared to single-frame retrieval methods. As a result, performing multi-frame place recognition can be extremely slow for deployment on autonomous vehicles or evaluation on large datasets, and fail when using relatively short parameter values such as a sequence length of 2 frames. In this paper, we propose DeepSeqSLAM: a trainable CNN+RNN architecture for \textit{jointly learning visual and positional representations} from a single monocular image sequence of a route. We demonstrate our approach on two large benchmark datasets, Nordland and Oxford RobotCar---recorded over 728 km and 10 km routes, respectively, each during 1 year with multiple seasons, weather, and lighting conditions. On Nordland, we compare our method to two state-of-the-art sequence-based methods across the entire route under summer-winter changes using a sequence length of 2 and show that our approach can get over 72\% AUC compared to 27\% AUC for Delta Descriptors and 2\% AUC for SeqSLAM; while drastically reducing the deployment time from around 1 hour to 1 minute against both. The framework code and video are available at \url{mchancan.github.io/deepseqslam}
\end{abstract}

\begin{figure}[!h]
  \centering
  \includegraphics[width=0.3\linewidth]{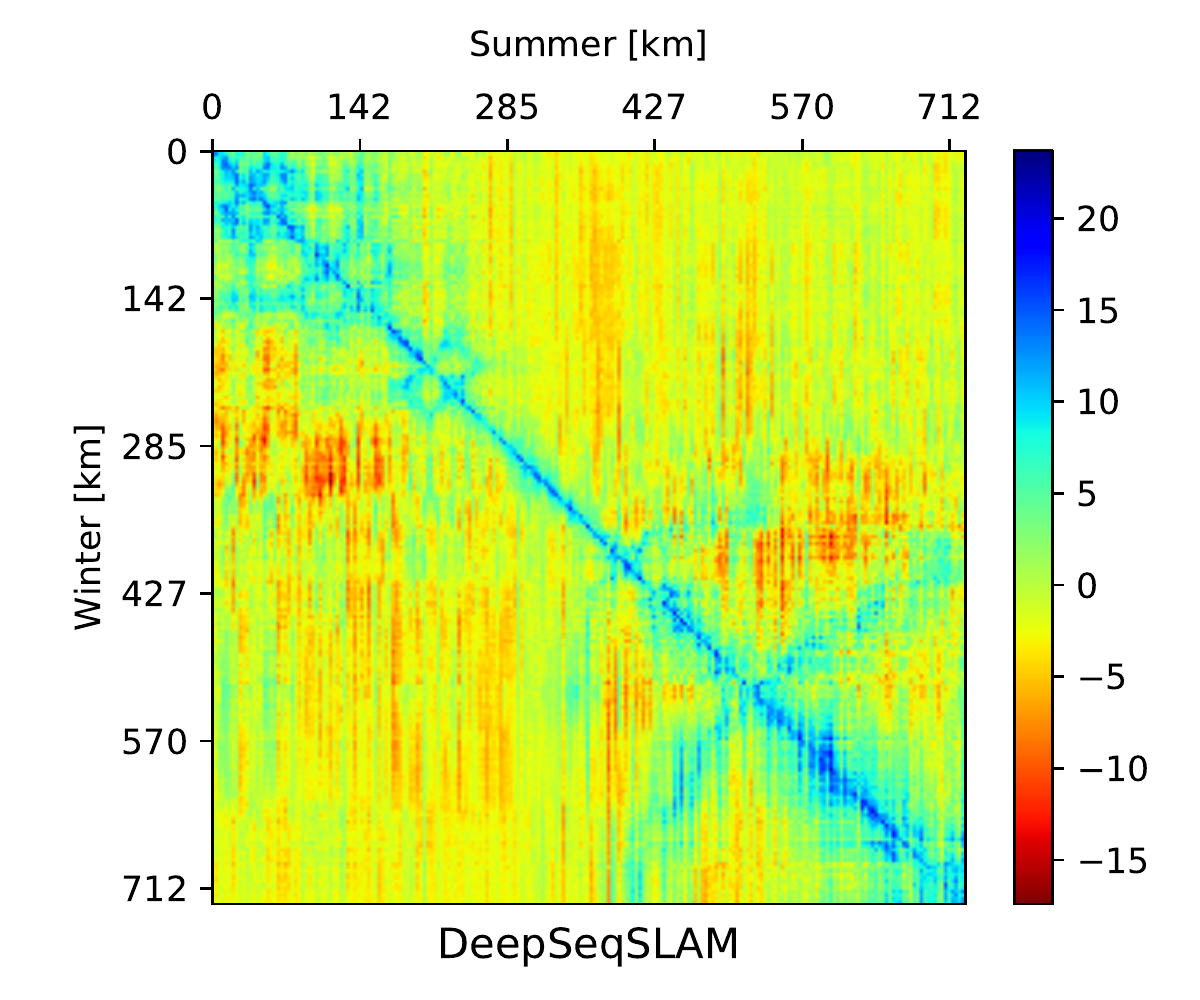}
  \includegraphics[width=0.3\linewidth]{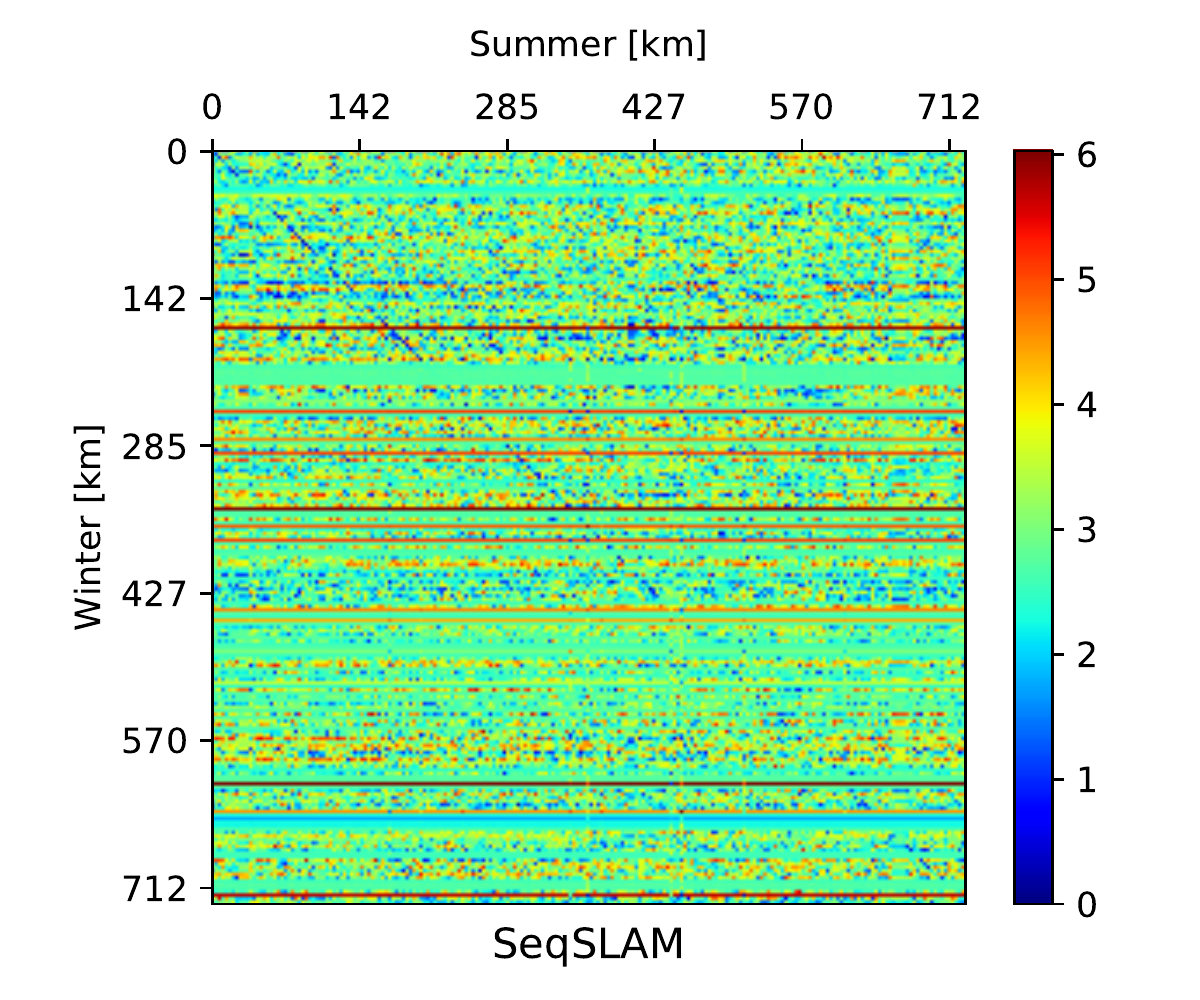}
  \includegraphics[width=0.3\linewidth]{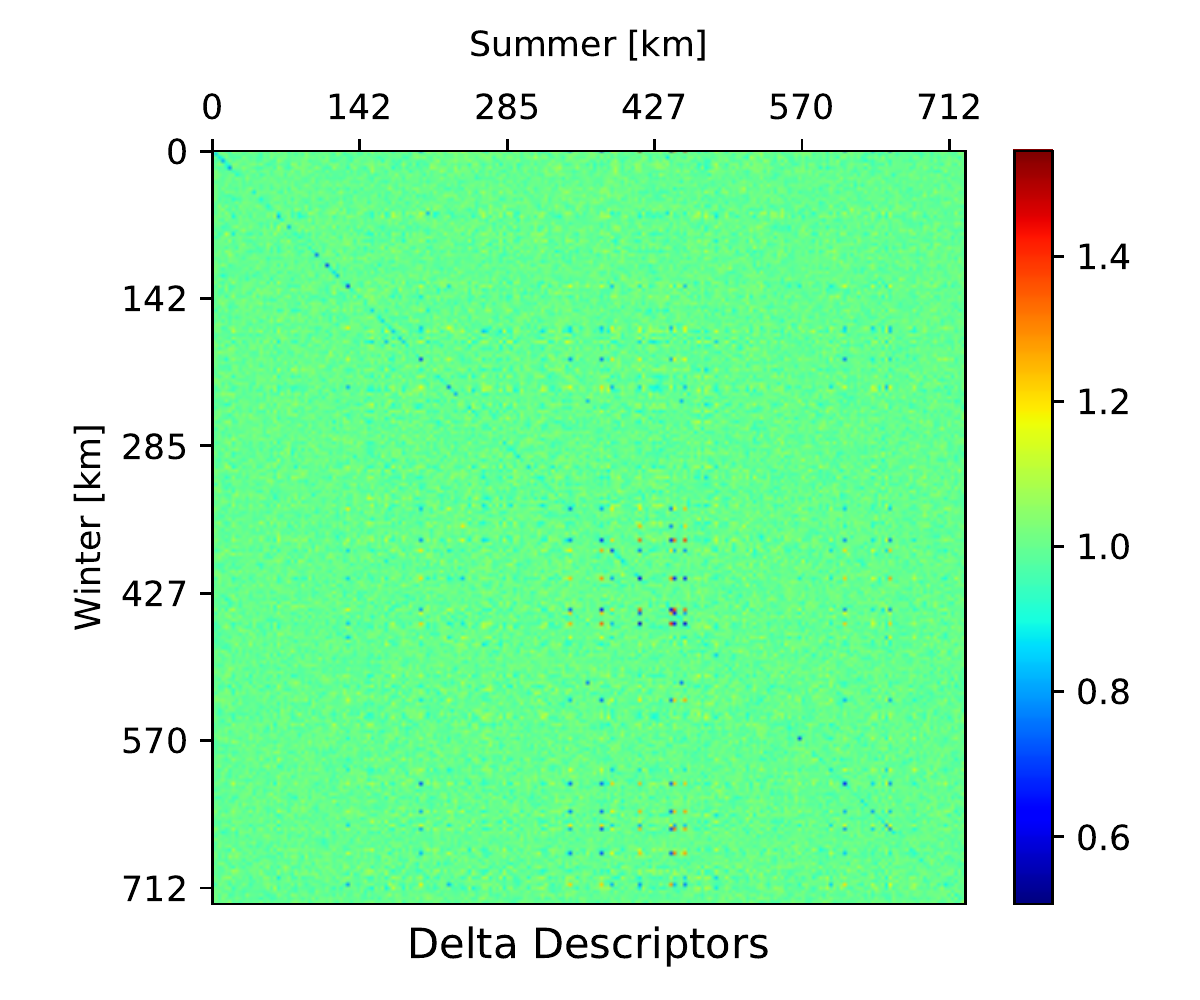}
  \caption{\textbf{State-of-the-art comparison of sequential matching between \textit{summer} (reference) and \textit{winter} (query) conditions on the Nordland dataset}. We use a challenging (short) sequence length of 2 frames to evaluate these methods on the full 728 km route of the dataset---comprising 3577 frames per reference/query traversal recorded at 0.1 FPS. \textbf{Left}: Neural activity profiles of the linear output layer of DeepSeqSLAM yielding over 72\% average precision (higher unit activation (\textit{blue}) represents a strong match candidate). \textbf{Middle, Right}: Difference matrices obtained using classical multi-frame methods such as SeqSLAM and Delta Descriptors resulting in 2\% and 27\% average precision, respectively (lower score (\textit{blue}) means a strong match candidate).}
  \label{sim_matrices}
\end{figure}

\section{Introduction and Related Work}

Localization is a key component for enabling the deployment of autonomous vehicles (AV) with full self-driving capabilities in real environments, including a range of other driver assistance features such as place recognition \citep{netvlad, scalevpr, vprsurvey,locationrecognition,convnetlandmarks}, multi-object tracking \citep{wang2019ad,chiu2020probabilistic}, road scene understanding \citep{barnes2017urban,autovision,3dperceptionforav}, obstacle detection \citep{obstacle,mcallister2017concrete}, and also large-scale driving datasets \citep{geiger2012kitti,robotcar,safetodrive,mapillary,yu2020bdd100k} for training autonomous driving (AD) models \citep{Xu_2017_CVPR,amini2019variational,amini2020end}. In place recognition research, there have been great advances on the use of convolutional neural networks (CNN) \citep{deeplearning} for visual place recognition \citep{netvlad,convnetperformace,scalevpr,pointnetvlad, attentionvpr} and visual localization tasks under extreme changing conditions \citep{d2net,benchmarking6dof}. These models have shown to generalize well across challenging appearance and viewpoint variations including multiple seasons, weather, and lighting conditions \citep{scalevpr,pointnetvlad,benchmarking6dof}. On the other hand, multi-frame-based methods for sequential matching are known for improving single-image retrieval results on driving datasets with significant appearance changes \citep{seqslam, vysotska2015lazy, vysotska2015efficient, naseer2018robust, toft2020long}. However, state-of-the-art sequence-based approaches like SeqSLAM \citep{seqslam} heavily rely on computationally expensive heuristics to further enhance the difference matrix between query and reference image sequences of the same route. Although researchers have proposed a number of sequence-based methods \citep{smart, deltad, 9109951}, these often exhibit the fundamental limitations of SeqSLAM: a) computational cost scale linearly with the dataset size, b) fail when using relatively short parameter values such as a sequence length of 2, and c) requiring both reference and query traversals to be pre-aligned with the same number of frames. 

In this work, we study the use of recurrent neural networks (RNN) \citep{lstm} for temporally integrating sequential information over short image sequences \citep{facil2019condition}; while leveraging the computational and dynamic information processing functions of these networks for end-to-end positional and sequence-based place learning for the first time. Hence, we propose a CNN+RNN architecture that can be trained end-to-end for \textit{jointly learning visual and positional representations}. Our approach also enables the use of any pre-trained CNN model, where the RNN component can then be trained using a single monocular traversal of a driving dataset. In contrast to classical sequence-based methods \citep{seqslam,smart,deltad, 9109951}---which often require velocity information or reference-query image sequences being of the same size---and recent neuro-inspired systems 
\citep{8870939,8756053,chancan2020hybrid}, our approach does not require velocity data and can work with query traversals of any size---only requiring positional information of the AV when traversing an environment, which can be obtained from visual odometry (VO) \citep{odometry4an}, radar odometry (RO) \citep{oxfordradar}, LiDAR, structure from motion (SfM) \citep{schoenberger2016sfm} or any other source of motion estimation. Thus enabling our approach to also work as a full SLAM system by encoding two key sensor modalities for autonomous navigation within a single \textit{motion and visual} perception framework \citep{chancan2020mvp}. We also show that our approach is orders of magnitude faster than classical methods and can be deployed on CPU, GPU, or multi-GPUs across many servers. 


\section{DeepSeqSLAM}
\label{gen_inst}

In robot navigation research for AD, both visual and positional information share the underlying representation that captures the nature of an environment as the robot navigates its surroundings. Contrary to the classical two-stage place recognition pipelines, i.e. match-then-temporally-filter, we propose to jointly perform \textit{visual and positional representation learning} to obtain bimodal descriptors that can be simultaneously used for performing sequence-based place recognition. We then refer to our approach as DeepSeqSLAM and detail its components in the next Section. 

\subsection{Global Place Description}

Given an image sequence $I_t$ of an environment, we apply a CNN function on each input image to obtain compact $n-$dimensional global image descriptors $\mathbf{d}_t$, where $n$ $\in$ $\mathbb{N}^+$ is a function of the CNN model. These representations can be learned through conventional backpropagation in a training stage, and then used to find the top-k similar images across a query sequence of the same route; potentially with different viewpoints and environmental conditions. After training, Cosine or Euclidean distances are often used to compute a similarity matrix between reference and query images.

\subsection{Sequence-based Place Learning}

Classical multi-frame-based place recognition methods are required to first compute a pairwise difference matrix between both reference and query image sequences to then iteratively apply temporal filtering/searching heuristics on top of these results to further reduce false-positive rates. These traditional methods have demonstrated to be highly inefficient and also impractical for deployment on real AVs or large datasets. Particularly when we want to deploy a localization system on a single query traversal after the training stage has been done on a reference route of the same environment.

\begin{table}[t]
  \vspace{-2mm}
  \caption{Two Large-Scale Benchmark Datasets for Sequence-based Place Recognition.}
  \vspace{-2mm}
  \label{table_dataset}
  \centering
  \begin{tabular}{lcccc}
    \toprule
    Dataset     & Environmental Condition     & Journey & Size (frames) & Sensor \\
    \midrule
     & Summer  & 728 km &  35768  &  \\
    \multirow{2}{*}{Nordland Railway}     & Fall & 728 km & 35768  & 1$\times$ camera   \\
         & Winter &   728 km     &  35768& (1 FPS) \\
         & Spring   &  728 km   &  35768 \\
    \midrule
     & Day  & 10 km & 31472 &  \\
    \multirow{2}{*}{Oxford RobotCar}     & Night & 10 km & 38315  & 1$\times$ camera   \\
         & Snow &   10 km    & 40943 & (15 FPS) \\
         & Rain   &   10 km  &  37564 & \\
    \bottomrule
  \end{tabular}
  \vspace{-5mm}
\end{table}

Here we propose to integrate an RNN model on top of a CNN function for sequence learning from a single traversal of a route. Depending on the CNN model, we can alleviate the large amount of data---typically required for effective training---by plugin a pre-trained CNN-based network such as NetVLAD \citep{netvlad}; heavily used in visual place recognition research for extracting robust global image descriptors. Thus, the CCN function can be freeze while the RNN component is trained. The required input for the RNN is a global image descriptor $\mathbf{d}_t$ (obtained from full-resolution RGB images) concatenated with its corresponding 2D global positional information $\textbf{p}_t$ (represented here using a 2-\textit{d} vector). In our experiments, we use the corresponding GPS data normalized between -1 and 1 for $\textbf{p}_t$, although other sources of motion estimation such as VO, radar. or LiDAR can be used.

\subsection{Implementation Details}

We use the NetVLAD \citep{netvlad} network (based on a VGG-16 architecture \citep{vgg} trained on Pittsburgh 30k \citep{7054472}) as our back-end CNN function for global image description, with a feature dimension of $n=4096$ and $l_2$-normalized. For the RNN component, a single cell, long short-term memory (LSTM) with 512 units is used, with a multi-layer perceptron (MLP) with $N$ units---equal to the number of frames in the training data---over the top for receiving the output state of the LSTM; which after training is capable of generating the neural activity profiles shown in Fig. \ref{sim_matrices}. We use a learning rate of 0.01 with Adam \citep{adam} for training. In sequence-based place recognition, the sequence length, $d_s$  $\in$ $\mathbb{N}^+$, is a sensitive parameter \citep{jake} that is typically predefined to arrange the training data in short consecutive overlapping sequences of size $d_s$ frames. We provide extensive experimental results on the trade-off of using different $d_s$ values from 1 to 24, and compare the performance of our method to two state-of-the-art multi-frame systems in the next Section.

\section{Experimental Evaluations}
\label{headings}

\subsection{Large-Scale Datasets}

In Table \ref{table_dataset} we report the details of the datasets used to demonstrate our approach. The Nordland dataset \citep{nordland} was recorded on a 728 km train journey in Norway, providing four long traversals, once per season, with diverse visual conditions over 1 year. We train our approach on a full traversal (\textit{summer}) and test on the remaining (\textit{fall}, \textit{winter}, \textit{spring}). The Oxford RobotCar dataset \citep{robotcar} was collected on a car platform over a 10 km route in Oxford, UK, also over 1 year. The full dataset includes a range of sensory information from LiDAR, monocular cameras, and trinocular stereo cameras, over +100 traversals with different weather, seasons, and dynamic urban conditions. We select 4 sequences of this dataset\footnote{Ref. as 2015-05-19-14-06-38, 2014-12-10-18-10-50, 2015-02-03-08-45-10, 2015-10-29-12-18-17 in [24].}and train our method on a single traversal.

\subsection{State-of-the-Art Results}

\begin{figure}[!t]
  \centering
  \includegraphics[width=0.32\linewidth]{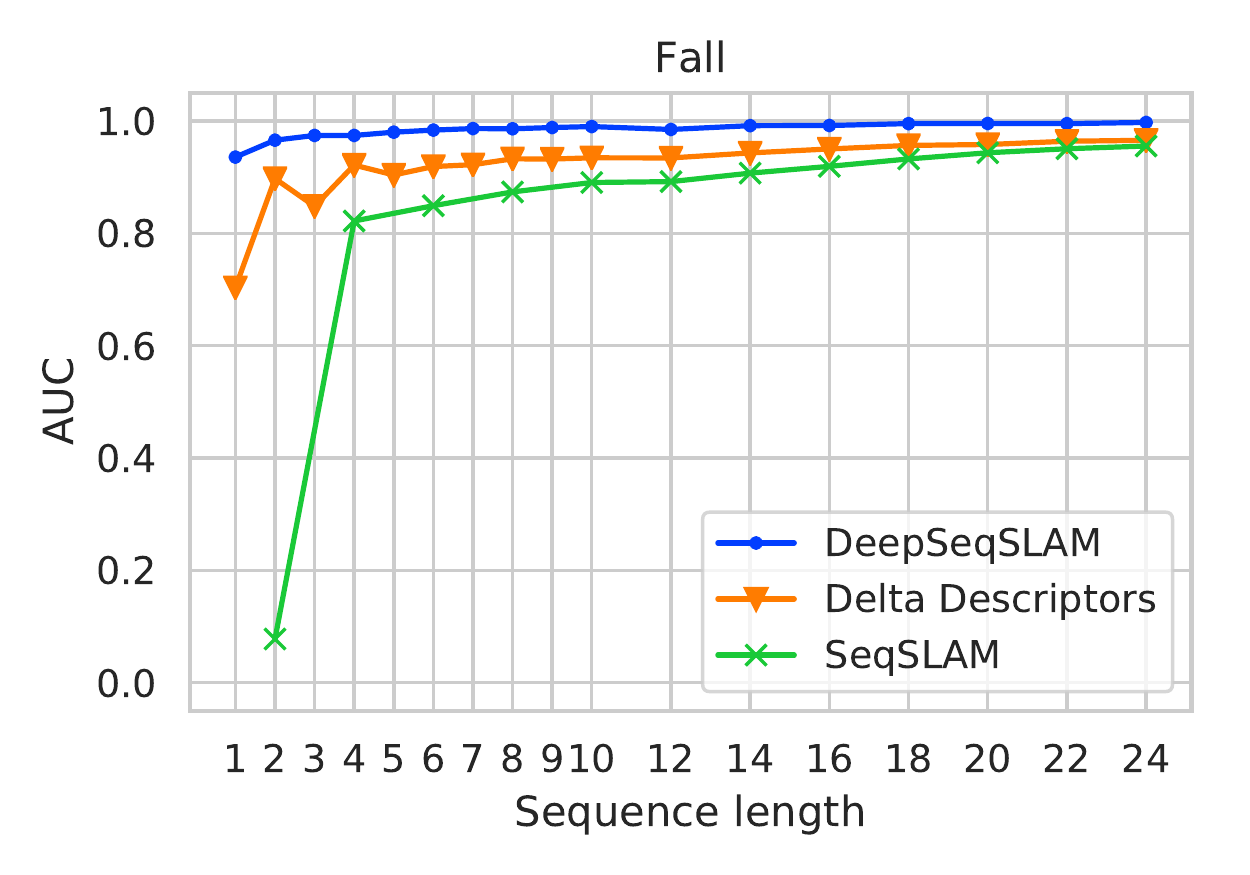}
  \includegraphics[width=0.32\linewidth]{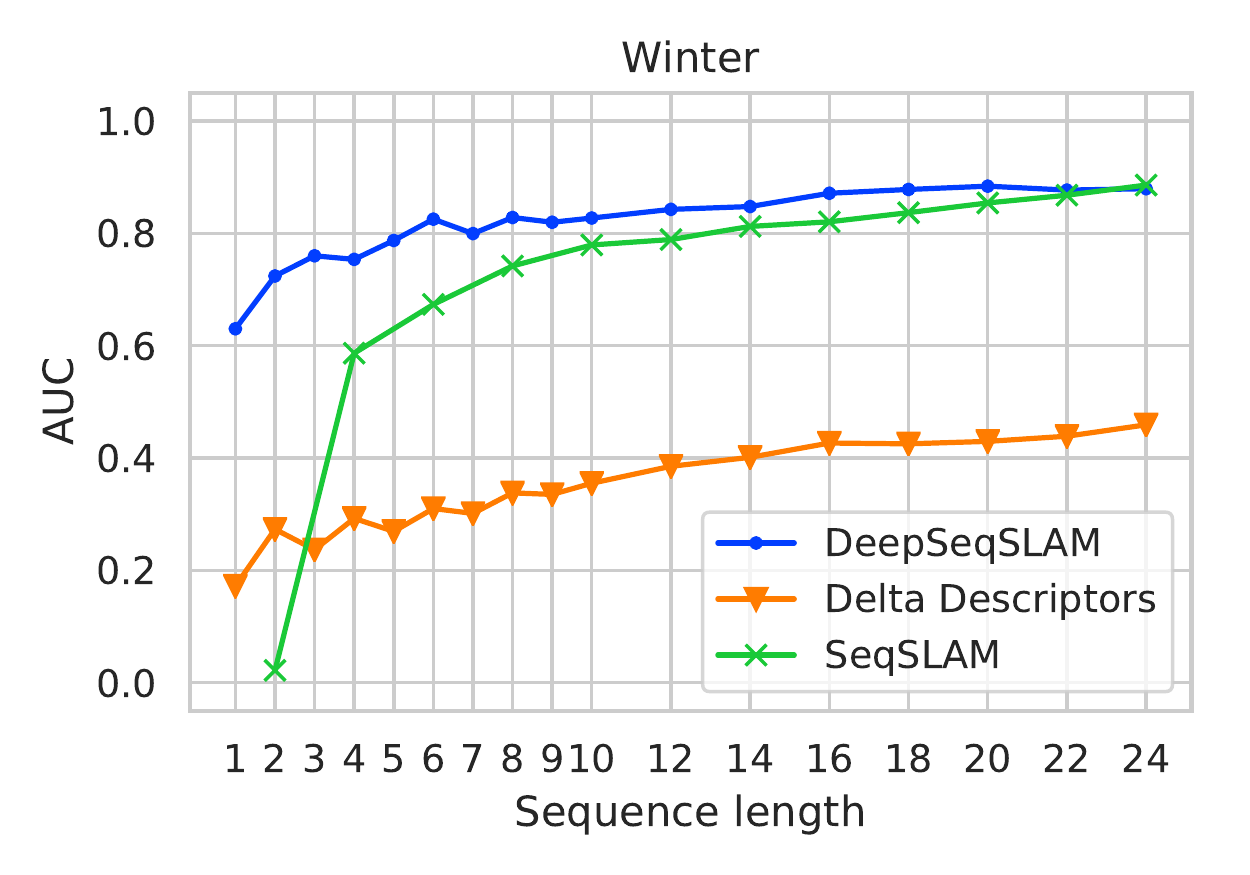}
  \includegraphics[width=0.32\linewidth]{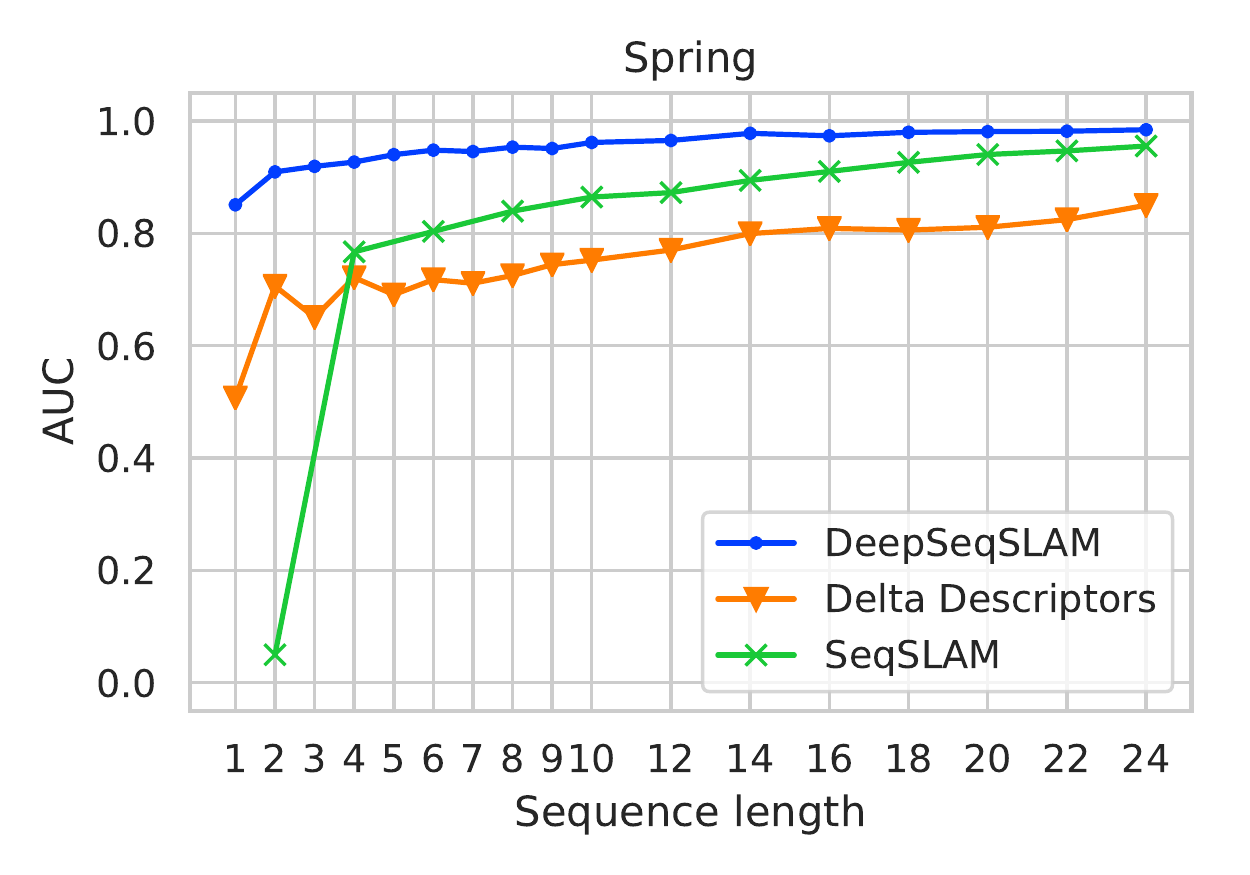}
  \vspace{-3mm}
  \caption{\textbf{Influence of sequence length} ($d_s$) in multi-frame methods and its generalization capabilities from reference (\textit{summer}) to query traversals (\textit{fall}, \textit{winter}, and \textit{spring}) of the Nordland dataset.}
  \label{ds}
\end{figure}

\begin{figure}[!t]
  \centering
  \includegraphics[width=0.32\linewidth]{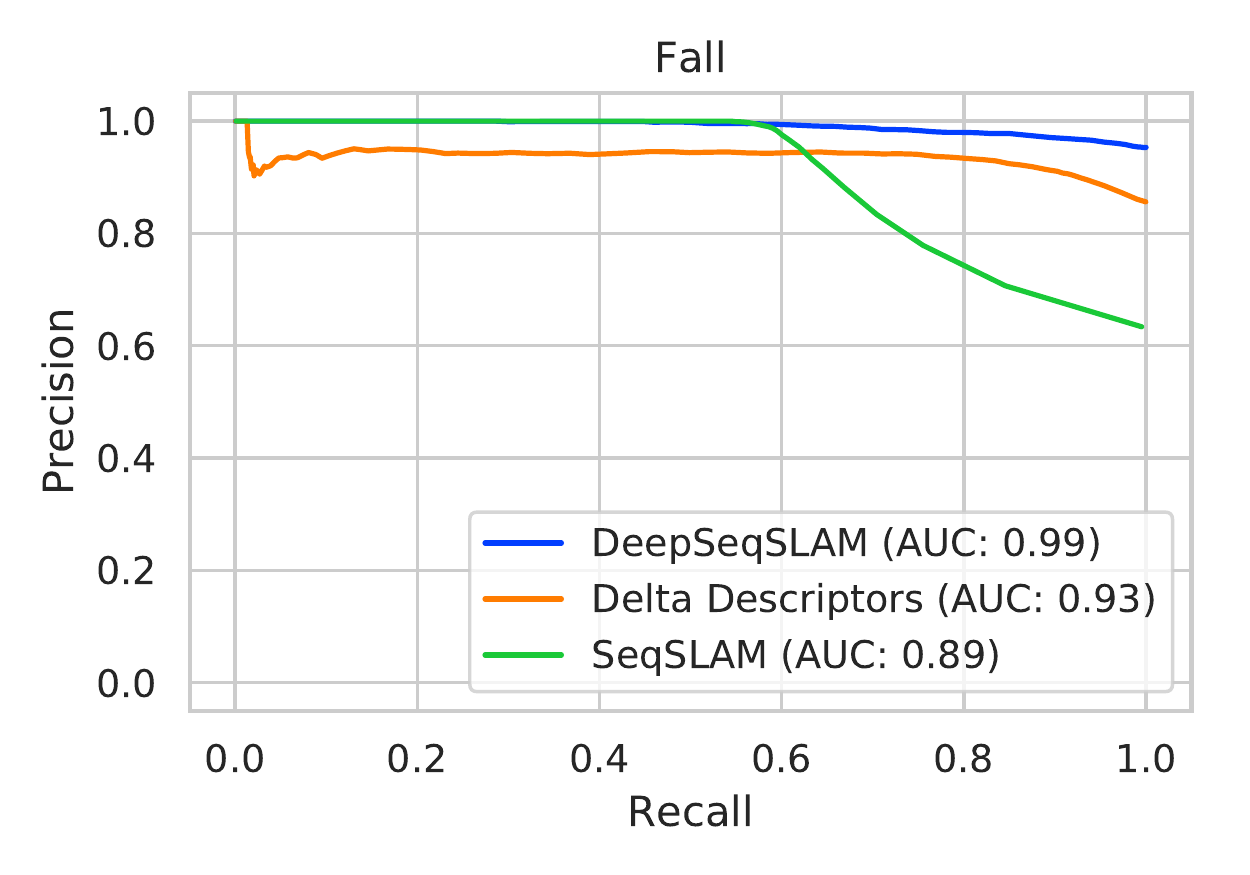}
  \includegraphics[width=0.32\linewidth]{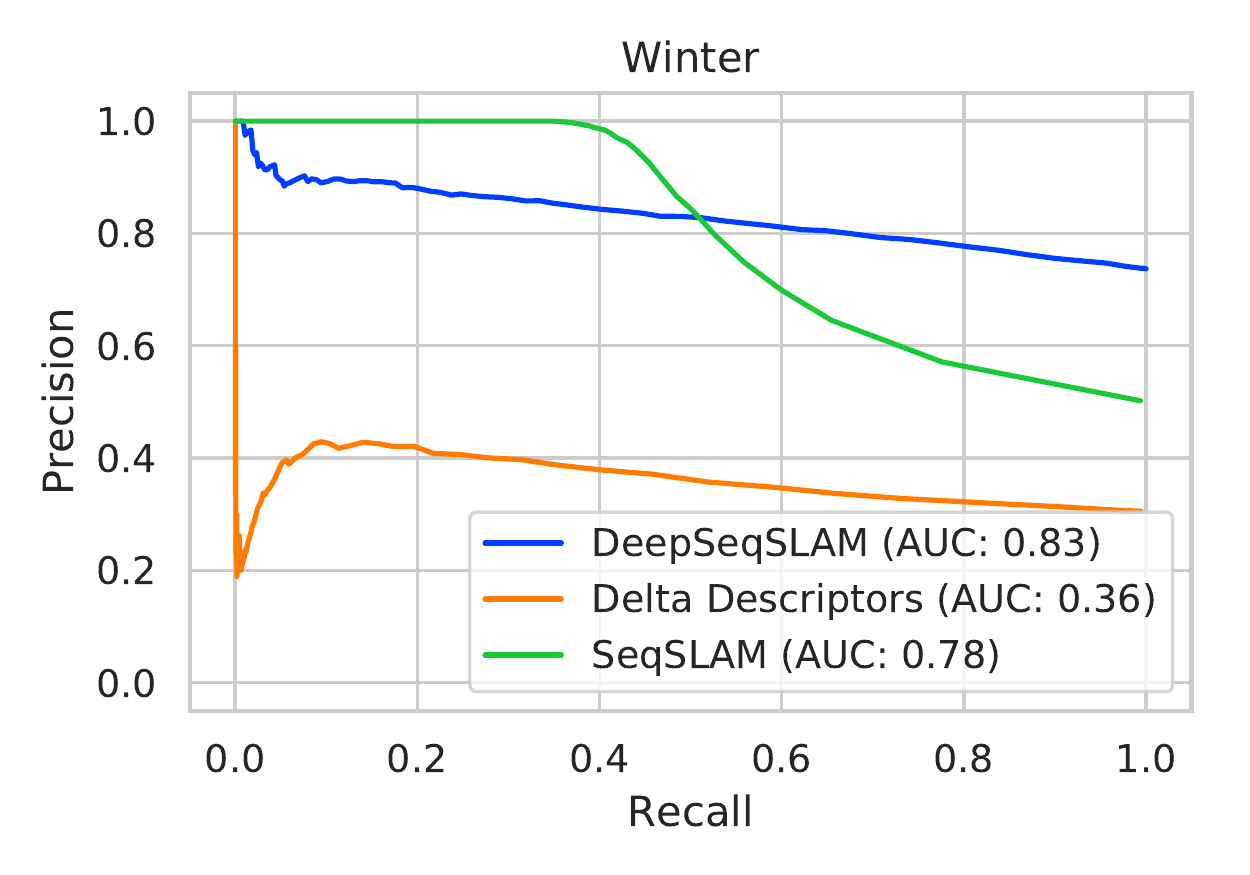}
  \includegraphics[width=0.32\linewidth]{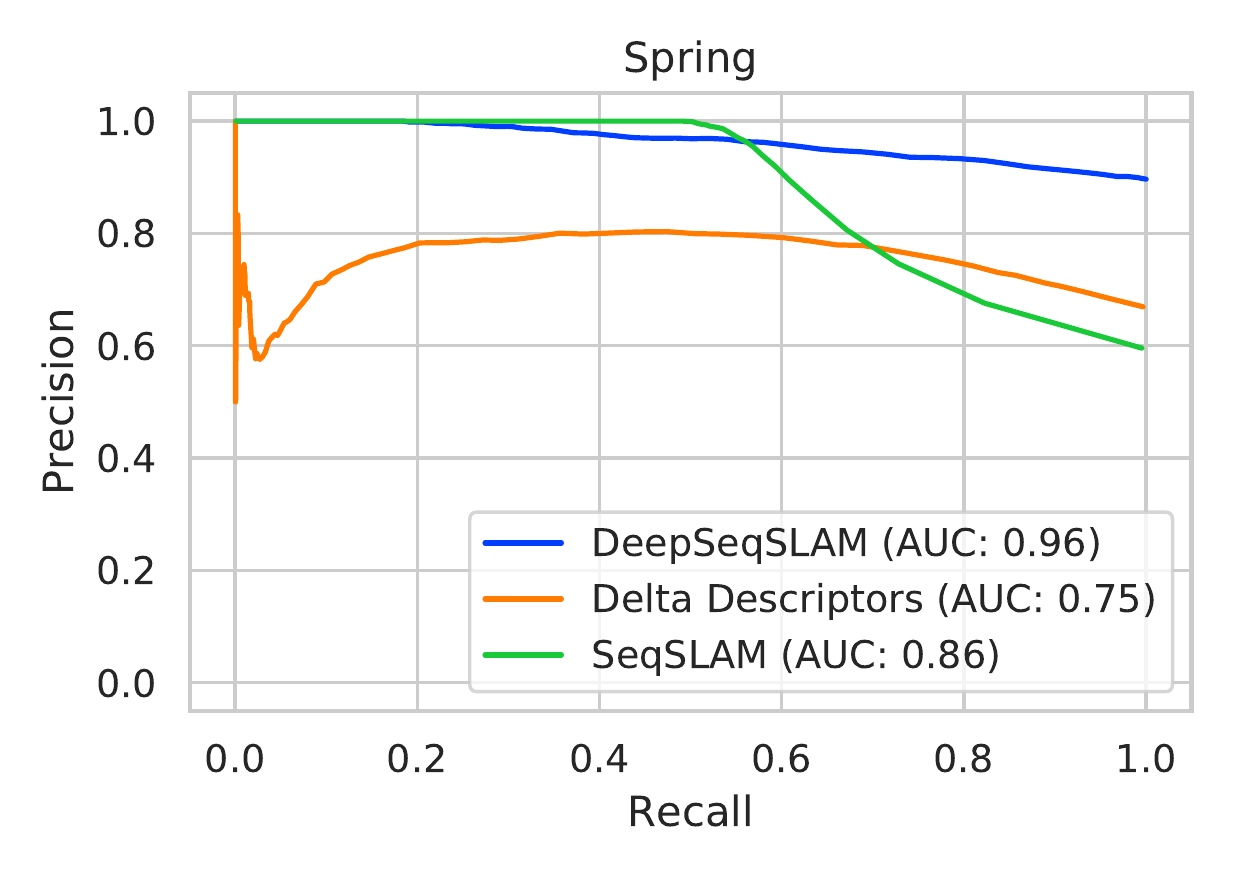}
  \vspace{-3mm}
  \caption{\textbf{State-of-the-Art comparison of PR performance on Nordland} with sequence length 10.} 
  \vspace{-4mm}
  \label{pr}
\end{figure}

In Fig. \ref{ds} we study the influence of changing the sequence length on our proposed method trained on the \textit{summer} and compare their generalization capabilities to \textit{fall}, \textit{winter}, and \textit{spring} conditions with classical sequence-based approaches such as SeqSLAM \citep{seqslam} and the recently published Delta Descriptors work \citep{deltad}. For SeqSLAM, we use the original MATLAB implementation provided in \citep{openseqslam2} with default parameters (only changing the sequence length $d_s$ as required); the SeqSLAM algorithm only works using an even number for $d_s$ though. For Delta Descriptors, we use the original Python code provided in \citep{deltad} with default parameters also varying the sequence length only as needed. We highlight that DeepSeqSLAM is getting such good performance with very short sequence lengths, compared to classical methods. In particular, under \textit{summer} to \textit{winter} changes with a sequence length of 2 (Fig. \ref{ds}-Middle), SeqSLAM and Delta Descriptors get 2\% and 27\% AUC, respectively, while our method gets over 72\% AUC. In Fig. \ref{sim_matrices} we provide the full similarity matrix results for each method under this challenging setting. Furthermore, in Fig. \ref{pr} we report the PR curves for a sequence length of 10 across all these models, including its corresponding AUC metrics. We note that SeqSLAM performs competitively at this $d_s$ with 78\% AUC on \textit{winter}, Fig. \ref{pr}-middle, but DeepSeqSLAM outperforms it with 83\% AUC. For computing the average-precision/AUC results from Figs. \ref{sim_matrices}, \ref{ds} and \ref{pr}, given a \textit{query} image $I_{q_t}$ and its corresponding \textit{reference} image $I_{d_t}$, the retrieved images between $I_{d_{t- \delta}}$ and $I_{d_{t+\delta}}$ were considered as a correct match; where the tolerance $\delta$ was given by $d_s+10$. An overview of the proposed CNN+RNN architecture is presented in Fig. \ref{model} of the \textbf{Appendix}, along with its training curves (accuracy and loss traces) in Fig. \ref{training}, Table \ref{table_time} on the computational performance analysis, and qualitative results on the Nordland dataset in Fig. \ref{qualy}.

\section{Discussion and Conclusions}
\label{others}

The results on the full Nordland dataset show that our proposed CNN+RNN model is capable of learning meaningful temporal relations from a single image sequence of a large driving dataset; while significantly outperforming classical sequence-based methods in runtime, accuracy, and computational requirements. We used a small two-layer CNN for exploring the end-to-end training behavior (from scratch) of DeepSeqSLAM but our preliminary results showed that the CNN component does not generalize well to drastic visual changes; which was expected since these models require a significant amount of data for effectively training and generalizing. We see this observation as future work to further investigate the advantages of jointly learning visual and positional information for AD applications. Finally, we call our method DeepSeqSLAM since it is our goal to build a full simultaneous localization and mapping (SLAM) system using this framework by incorporating a learning-based component for geometric mapping such as those in \citep{mapnet2018,bian2019depth,zhao2020towards}. Our first attempts to integrate these mapping systems into our framework were limited by different, expensive training requirements from these models, but we found that on the deployment stage our approach can be easily integrated with a mapping system and we are currently exploring this for future work.


\small
\bibliography{refs}







\newpage

\appendix

\section{DeepSeqSLAM architecture}

\begin{figure}[!h]
  \centering
  \includegraphics[width=0.54\linewidth]{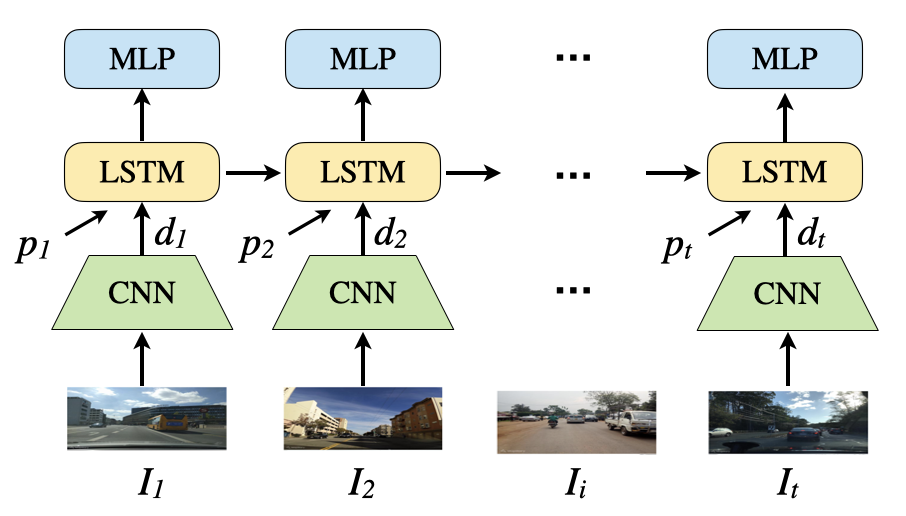}
  \caption{Proposed CNN+RNN architecture for learning positional $\mathbf{p}_t$ and visual $\mathbf{d}_t$ representations.}
  \label{model}
\end{figure}

\section{Training curves}

\begin{figure}[!h]
  \centering
  \includegraphics[width=0.4\linewidth]{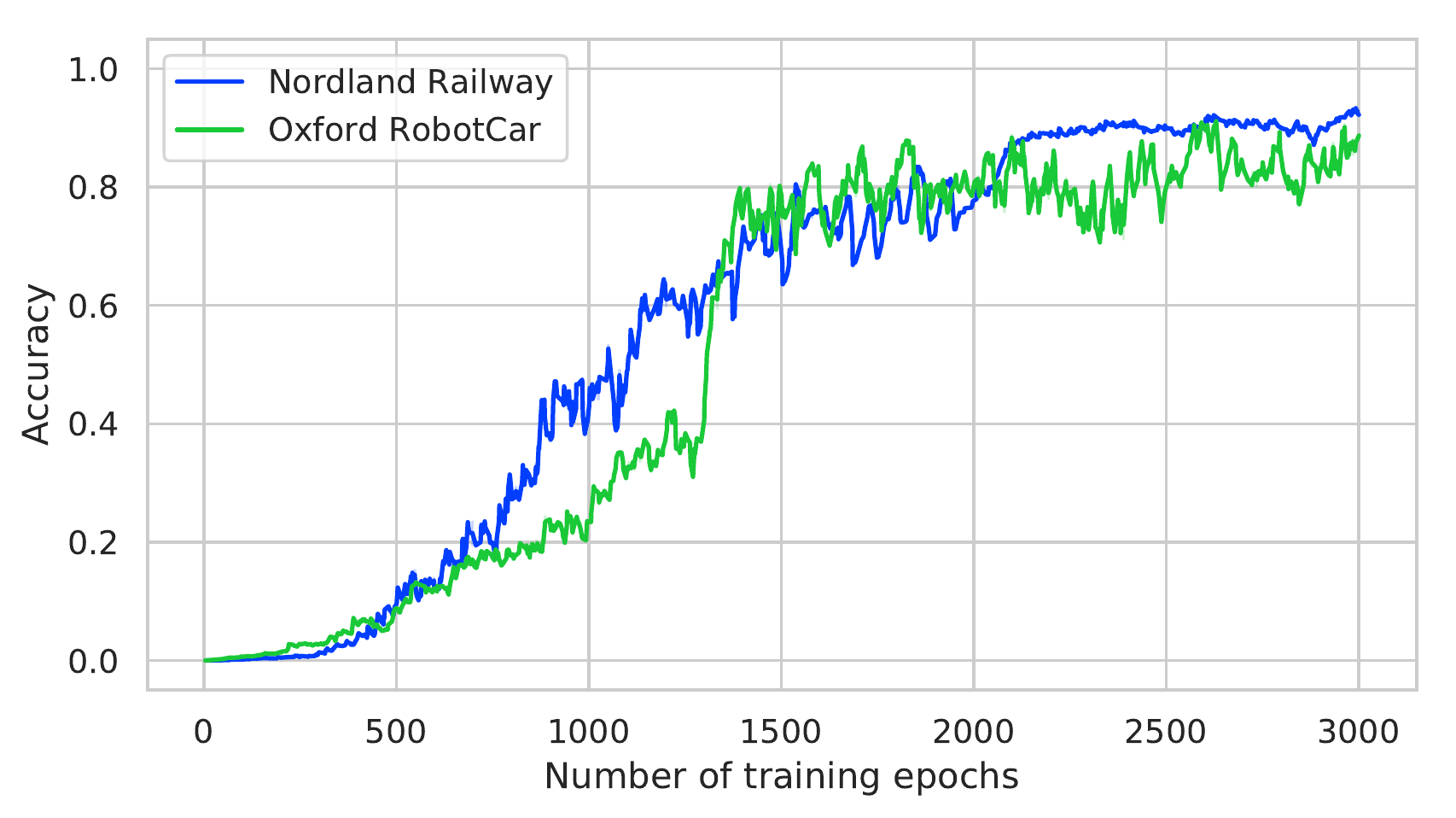}
  \includegraphics[width=0.4\linewidth]{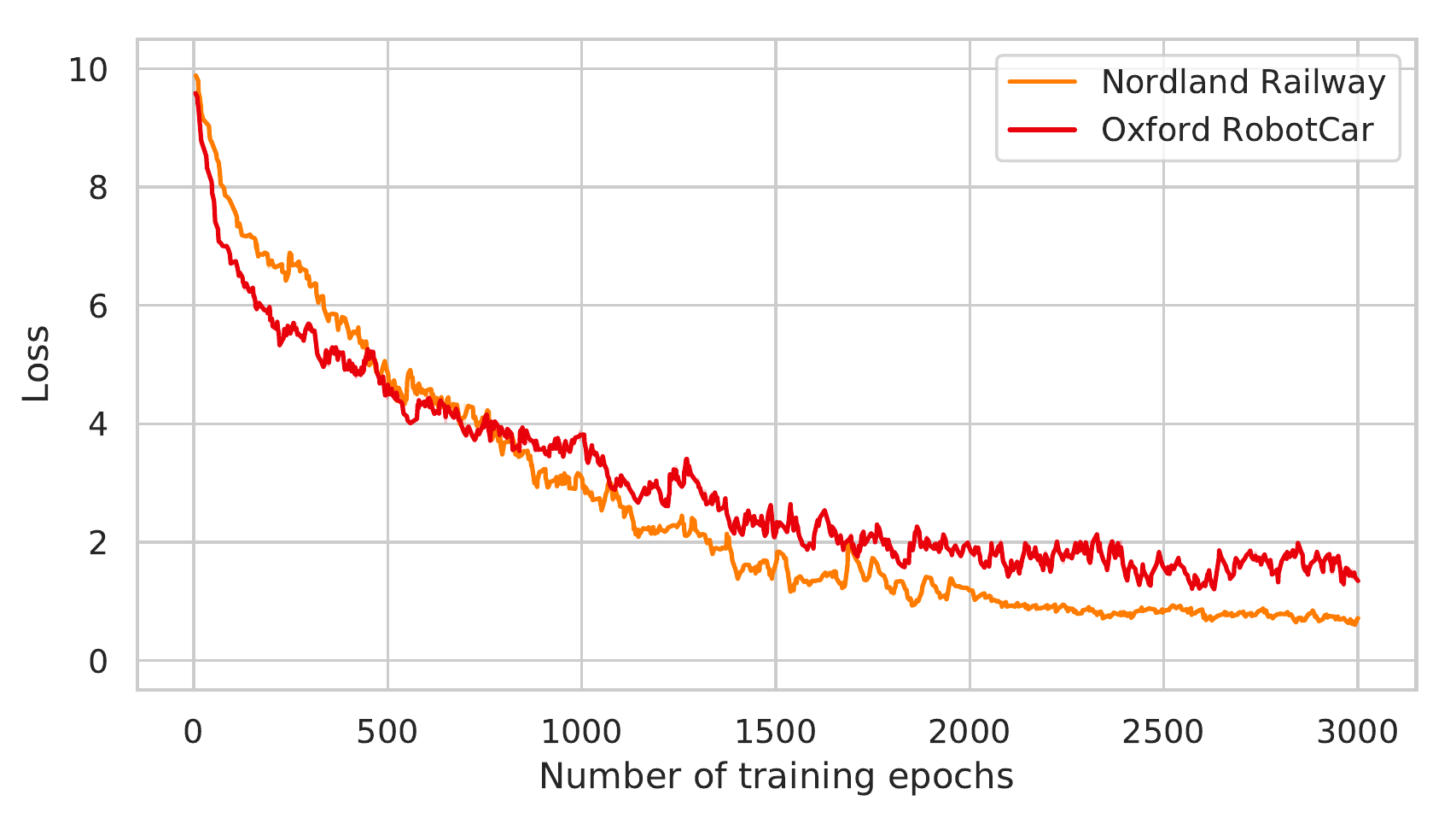}
  \caption{Training curves of DeepSeqSLAM on a single entire traversal of the Nordland Railway and the Oxford RobotCar datasets.}
  \label{training}
\end{figure}

In Fig. \ref{training} we present the training curves of our proposed approach on a full traversal of the Nordland and Oxford dataset; \textit{summer} and \textit{day}, respectively. Our results and state-of-the-art comparison on the Oxford dataset are currently in preparation, although the SeqSLAM algorithm does not achieve state-of-the-art performance on this dataset due to significant viewpoint changes and dynamic objects. While in our system, the CNN component makes our method robust to viewpoint variations.

\section{Computational Performance Analysis}

In contrast to traditional methods, that can only run on CPU and are computationally expensive on large datasets, our PyTorch \citep{NIPS2019_9015} implementation of DeepSeqSLAM has been designed to run on CPU, GPU, and multi-GPUs across many servers. We plan to evaluate DeepSeqSLAM on much larger datasets and also use it for training novel CNN architectures that may require preprocessing stages such as image augmentation for instance. The full source-code of DeepSeqSLAM is made publicly available for the benefit of the community.

\begin{table}[!h]
  \caption{Computational efficiency on deployment}
  \label{table_time}
  \centering
  \begin{tabular}{lccc}
    \toprule
    & \multicolumn{3}{c}{Runtime on a full query traversal of the Nordland dataset}                   \\
    \cmidrule(r){2-4}
    Method      & Sequential Matching & Size (frames) & Device \\
    \midrule
    \textbf{DeepSeqSLAM}*   & \textbf{1 min} & \textbf{35768}  & \textbf{CPU} \\
    SeqSLAM      & 70 min & 35768  & CPU   \\
    Delta Descriptors  & 51 min      & 35768 & CPU \\
    \bottomrule
  \end{tabular}
  \\ \footnotesize{*DeepSeqSLAM can also run on a single GPU or even multiple GPUs across many servers.}
\end{table}

\section{Qualitative Results on the Nordland Railway Dataset}

\begin{figure}[!t]
  \centering
  \includegraphics[width=0.9\linewidth]{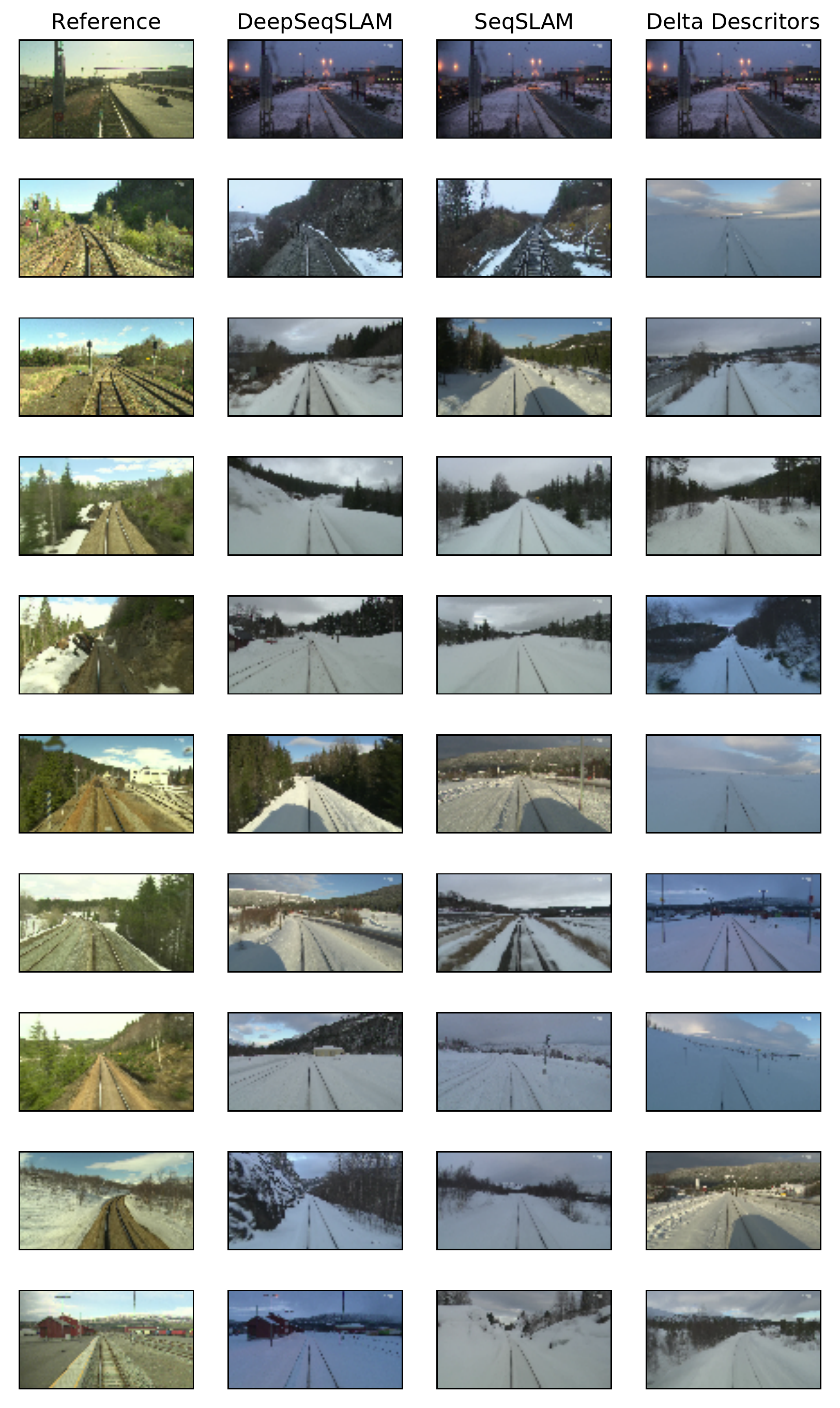}
  \caption{\textbf{Qualitative results on the full 728km route of the Nordland Railway dataset}. The reference and query image sequences are \textit{summer} and \textit{winter}, respectively, for each method.
  }
  \label{qualy}
\end{figure}

\end{document}